# Accuracy improvement of robot-based milling using an enhanced manipulator model

**Alexandr Klimchik**[a,b], **Yier Wu**[a,b], **Stéphane Caro**[b,c], **Benoît Furet**[b,d], **Anatol Pashkevich**[a,b]

[a] Ecole des Mines de Nantes, France
[b] Institut de Recherche en Communications et Cybernétique de Nantes, France
[c] French National Centre for Scientific Research, Nantes, France
[d] University of Nantes, Nantes, France
e-mail: alexandr.klimchik@emn.fr, yier.wu@emn.fr, stephane.caro@irccyn.ec-nantes.fr, benoit.furet@univ-nantes.fr, anatol.pashkevich@emn.fr,

**Abstract.** The paper is devoted to the accuracy improvement of robot-based milling by using an enhanced manipulator model that takes into account both geometric and elastostatic factors. Particular attention is paid to the model parameters identification accuracy. In contrast to other works, the proposed approach takes into account impact of the gravity compensator and link weights on the manipulator elastostatic properties. In order to improve the identification accuracy, the industry oriented performance measure is used to define optimal measurement configurations and an enhanced partial pose measurement method is applied for the identification of the model parameters. The advantages of the developed approach are confirmed by experimental results that deal with the elastostatic calibration of a heavy industrial robot used for milling. The achieved accuracy improvement factor is about 2.4.

## 1   Introduction

At present, the conventional CNC machines are progressively replaced in industry by robotic manipulators to perform main manufacturing tasks. For those applications, industrial robots are considered to be very competitive due to their manufacturing flexibility, large workspace and cost-effectiveness. At the same time, the robotic-based machining introduces some difficulties. For instance, link and joint compliances become non-negligible when robot is under substantial external loading. So, in order to achieve high processing accuracy, essential revision of relevant mathematical models and control strategies are required.

The stiffness modeling of robotic manipulators has been in the focus of the research community for more than 30 years (Salisbury 1980). There exist different approaches that are able to take into account particularities of serial and parallel manipulators (Merlet and Gosselin 2008 , Kövecses and Angeles 2007). Among a number of existing stiffness modeling approaches, the Virtual Joint Modeling (VJM) method looks the most attractive in robotics. Its main idea is to take into account the elastostatic properties of flexible components by presenting them as equivalent localized virtual springs (Pashkevich et al. 2011). However the stiffness modeling of the manipulators with gravity compensators has not found enough attention yet. Another difficulty related to the stiffness modeling of robotic manipulators is the identification of their model parameters. This issue is quite new in robotics, the existing approaches are usually suitable for strictly serial manipulators only (Dumas et al. 2011). Therefore, this paper aims to obtain a sophisticated elasto-static model for heavy industrial robots with a gravity compensator and to identify their parameters.

## 2    Problem of the compliance errors compensation

In common engineering practice, robot behavior under an external loading can be described by the following force-deflection relation (Klimchik et al. 2012a)

$$\Delta \mathbf{t} = \left( \mathbf{J}_\theta \cdot \left( \mathbf{K}_\theta - \mathbf{H}_{\theta\theta} \right)^{-1} \cdot \mathbf{J}_\theta^T \right) \cdot \mathbf{F} \qquad (1)$$

where $\mathbf{J}_\theta$ and $\mathbf{H}_{\theta\theta}$ are the Jacobian and Hessian matrices respectively, the matrix $\mathbf{K}_\theta$ describes the elastic properties of the manipulator components. This model allows us to compute the end-effector deflection $\Delta \mathbf{t}$ due to the external loading $\mathbf{F}$. Since the manipulator deflection caused by the loading is known, it can be used to improve the positioning accuracy by means of error compensation technique (Fig. 1). However in practice, only geometrical parameters are provided by the robot manufacturer, while elastostatic parameters should be identified using dedicated calibration techniques. Usually the force-deflection relation (1) is rearranged in the linear model suitable for the identification procedure, which is a linear mapping between the parameters to be identified and the end-effector displacement

$$\Delta \mathbf{t}_i = \mathbf{A}_i \, \mathbf{k}; \qquad \mathbf{A}_i = \left[ \mathbf{J}_{1i} \mathbf{J}_{1i}^T \mathbf{F}_i, ..., \mathbf{J}_{ni} \mathbf{J}_{ni}^T \mathbf{F}_i \right] \quad (i = \overline{1, m}) \qquad (2)$$

where the vector $\mathbf{k}$ collects elastostatic parameters of the matrix $\mathbf{k}_\theta = \mathbf{K}_\theta^{-1}$.

It should be mentioned that such a model can be efficiently applied for strictly serial manipulators (without closed-loops) while for heavy manipulators with a gravity compensator this procedure should be revised in order to take into account particularities of the stiffness model. Another difficulty is related to the gravity compensator modeling, whose parameters are usually not given.

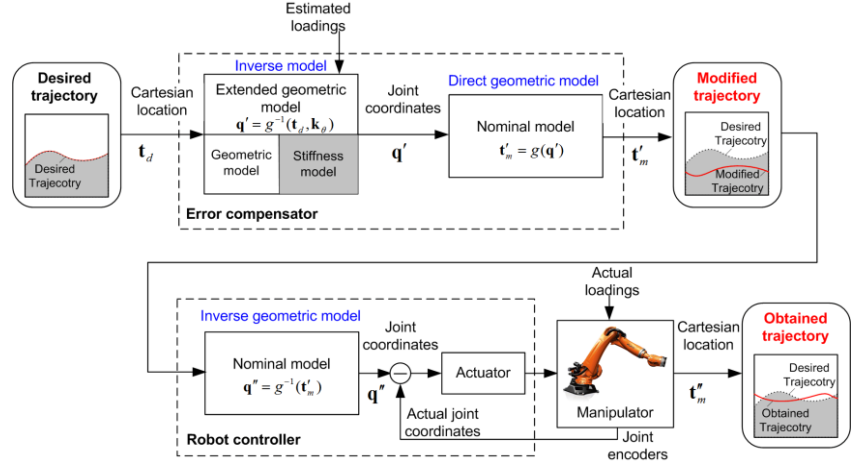

**Fig. 1** Off-line compliance errors compensation strategy

Hence, the goal of this work is to obtain a sophisticated elastostatic model that can be used for compliance errors compensation. Accordingly, two problems should be considered: (i) developing the model for the compensator and methodology for the identification of its parameters; (ii) integration of the compensator into conventional elastostatic model and identification of its parameters.

## 3   Parameters of the enhanced manipulator model and their identification

Considered industrial robot KUKA KR-270 incorporates gravity compensator that is used to balance link-weights but also affects manipulator elastostatic properties. The mechanical structure of the gravity compensator under study is presented in Fig. 2. The compensator incorporates a passive spring attached to the first and second links, which creates a closed loop that generates the torque applied to the second joint of the manipulator. The compensator geometrical model includes three node points $P_0$, $P_1$, $P_2$, which yield three principal geometrical parameters $L = |P_1, P_2|$, $a = |P_0, P_2|$, $s = |P_0, P_1|$. Let us also introduce some auxiliary parameters (such as $a_x$ and $a_y$), whose geometrical meanings are described in Fig. 2. The fact that the gravity compensator affects on the second joint only allows us to replace the constant parameter $K_{\theta_2}$ in the model (1) by the non-linear one that also takes into account elasto-static properties of the compensator.

The variable $s$ describing the compensator spring deflection can be computed as a function of the second joint coordinate $q_2$ as follows:

$$s^2 = a^2 + L^2 + 2\,a\,L\,\cos(\alpha - q_2) \qquad (3)$$

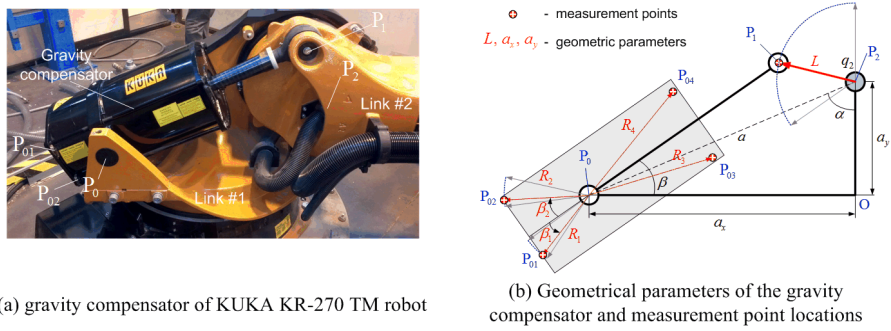

(a) gravity compensator of KUKA KR-270 TM robot

(b) Geometrical parameters of the gravity compensator and measurement point locations

**Fig. 2** Gravity compensator and its model

Therefore, the equivalent stiffness of the second joint (comprising both the manipulator and compensator stiffnesses) can be expressed as

$$K_{\theta_2} = K_{\theta_2}^0 + K_c\, a\, L \left( \frac{s_0}{s} \left( \frac{a\,L}{s^2} \sin^2(\alpha - q_2) + \cos(\alpha - q_2) \right) - \cos(\alpha - q_2) \right) \quad (4)$$

where $K_c$ is the gravity compensator stiffness, the value $s_0$ corresponds to the distance $|P_0, P_1|$ for the unloaded spring. This allows us to extend the classical stiffness model (1) of the serial manipulator by modifying the virtual spring parameters in accordance with the compensator properties. In this case, the Cartesian stiffness matrix $\mathbf{K}_C$ can be computed using the following expression:

$$\mathbf{K}_C = \left( \mathbf{J}_\theta \cdot (\mathbf{K}_\theta(\mathbf{q}) - \mathbf{H}_{\theta\theta})^{-1} \cdot \mathbf{J}_\theta^T \right)^{-1} \quad (5)$$

which includes both the first and second order derivatives (Jacobians and Hessians) of the functions $\mathbf{g}(\mathbf{q},\boldsymbol{\theta})$ describing the manipulator geometry (Pashkevich et al. 2011). Here, the vectors $\mathbf{q}$ and $\boldsymbol{\theta}$ collect actuator coordinates and the corresponding deflections.

The equivalent stiffness of the second joint (4) depends on several geometrical parameters ($L, a_x, a_y$) that are unknown and should be identified using reference points shown in Fig. 2b. By considering particularities of the experimental setup for the geometric parameters identification, where for each measurement of the point $P_1$ joint coordinate $q_2$ is given, the value of $L$ can be computed as

$$L = \sum_{i=1}^{m} \widehat{\mathbf{p}}_i^T \mathbf{R}\, \widehat{\mathbf{u}}_i \Big/ \sum_{i=1}^{m} \widehat{\mathbf{u}}_i^T \widehat{\mathbf{u}}_i \quad (6)$$

where $\widehat{\mathbf{p}}_i = \mathbf{p}_i - m^{-1} \sum_{i=1}^{m} \mathbf{p}_i$, $\widehat{\mathbf{u}}_i = \mathbf{u}_i - m^{-1} \sum_{i=1}^{m} \mathbf{u}_i$, $\mathbf{u}_i = [\cos q_i,\ \sin q_i,\ 0]^T$, $\mathbf{p}_i$ is the Cartesian coordinate vector of point $P_1$ for the $i$th measurement and $m$ is the number of measurements and the orthogonal matrix $\mathbf{R} = \mathbf{V}\mathbf{U}^T$ can be obtained using the following SVD-factorization $\sum_{i=1}^{m} \widehat{\mathbf{u}}_i \widehat{\mathbf{p}}_i = \mathbf{U}\, \boldsymbol{\Sigma}\, \mathbf{V}^T$. The remaining geometrical parameters ($a_x$ and $a_y$) are $x$ and $y$ coordinates of the vector

$$\mathbf{p}_0 = \frac{1}{2} \left( \mathbf{I} - \mathbf{n}\mathbf{n}^T \right) \left( \sum_{j=1}^{k} \sum_{i=1}^{m} \widehat{\mathbf{p}}_i^j \widehat{\mathbf{p}}_i^{jT} \right)^{-1} \sum_{j=1}^{k} \sum_{i=1}^{m} \widehat{s}_i^j \widehat{\mathbf{p}}_i^j + k^{-1} m^{-1} \mathbf{n}\mathbf{n}^T \sum_{j=1}^{k} \sum_{i=1}^{m} \widehat{\mathbf{p}}_i^j \quad (7)$$

where $\hat{\mathbf{p}}_i^j = \mathbf{p}_i^j - m^{-1}\sum_{l=1}^{m}\mathbf{p}_l^j$, $\hat{s}_i^j = \mathbf{p}_i^{jT}\mathbf{p}_i^j - m^{-1}\sum_{l=1}^{m}\mathbf{p}_l^{jT}\mathbf{p}_i^j$, $\mathbf{p}_i^j$ is the Cartesian coordinate vector of point $P_{0j}$ for the $i$th measurement, $k$ is the number of reference points and $m$ is the number of measurements. Here, the vector $\mathbf{n}$ is the last column of the matrix $\mathbf{V}$ of the following SVD-factorization $\sum_{j=1}^{k}\sum_{i=1}^{m}\hat{\mathbf{p}}_i^j\hat{\mathbf{p}}_i^{jT} = \mathbf{U}\,\Sigma\,\mathbf{V}^T$.

Since all geometrical parameters are known, the elastostatic ones can be identified. To take into account the compensator influence while retaining the approach developed for serial robots without compensators, manipulator elastostatic parameters can be identified into two steps. The first step aims to compute the extended set of elastic parameters that includes all equivalent virtual springs for the second joint by using the standard least-square technique

$$\mathbf{k} = \left(\sum_{i=1}^{m}\mathbf{B}_i^{(p)T}\mathbf{B}_i^{(p)}\right)^{-1}\cdot\left(\sum_{i=1}^{m}\mathbf{B}_i^{(p)T}\Delta\mathbf{p}_i\right) \qquad (8)$$

where the vector $\Delta\mathbf{p}_i$ is the small displacement of the end-effector under the external loading $\mathbf{F}_i$, matrix $\mathbf{B}_i^{(p)}$ is a rearranged matrix $\mathbf{A}_i$ that integrates positional components only and considers the shape and meaning of vector $\mathbf{k}$. The second step deals with the identification of the gravity compensator parameters and compliance of joint #2 that can be obtained from the following equation

$$\begin{bmatrix} K_{\theta_2}^0 & K_c & s_0\cdot K_c \end{bmatrix}^T = \left(\sum_{i=1}^{m_q}\mathbf{C}_i^T\mathbf{C}_i\right)^{-1}\left(\sum_{i=1}^{m_q}\mathbf{C}_i^T K_{\theta_{2i}}\right) \qquad (9)$$

where $m_q$ is the number of different angles $q_2$ in the experimental data,

$$\mathbf{C}_i = \begin{bmatrix} 1 & -a\,L\cdot\cos(\alpha - q_{2i}) & a\,L/s\cdot(a\,L/s^2\cdot\sin^2(\alpha - q_{2i}) + \cos(\alpha - q_{2i})) \end{bmatrix} \qquad (10)$$

In order to ensure high calibration efficiency, the design of experiments should be considered while choosing measurement configurations. To the best of our knowledge, the best results for particular industrial applications can be achieved by using the test-pose based approach (Klimchik et al. 2012b), which reduces optimal pose selection to the following optimization problem:

$$\mathrm{trace}\left(\mathbf{A}_0^{(p)}\sum_{j=1}^{m_q}\left(\sum_{i=1}^{m}\mathbf{A}_i^{j(p)T}\mathbf{A}_i^{j(p)}\right)^{-1}\mathbf{A}_0^{(p)T}\right) \to \min_{\{\mathbf{q}_i,\mathbf{w}_i\}} \qquad (10)$$

Here matrix $\mathbf{A}_0^{(p)}$ has the same structure as matrix $\mathbf{A}_i^{(p)}$, but is defined by the desired test pose configuration $\mathbf{q}_0$ and the external loading $\mathbf{F}_0$. The values of $\mathbf{q}_0$, $\mathbf{F}_0$ are usually related to a typical machining configuration and force generated by the tool-workpiece interaction. Such an approach allows us to ensure the highest positioning accuracy after compensation compliance errors caused by the technological process.

Using theoretical results presented in this section, it is possible to obtain a sophisticated elasto-static model that can be used for further error compensation. In the next section, these results are used to obtain the stiffness model of the KUKA KR-270 robot.

## 4 Experimental results and comparison analysis

The main geometric parameters of the gravity compensator are $L$, $a_x$ and $a_y$ (see Figure 2). They can be identified by using relative locations of points $P_0$ and $P_1$ with respect to point $P_2$. Since the adopted geometric model is a planar one, here the laser tracker base frame is defined in a particular way in order to ensure that the marker locations relative to the XY-plane are not significant. Another important issue is related to the selection of the marker point locations on the rigid part of the gravity compensator. To ensure high identification accuracy, these markers should be located on the opposite sides of the compensator rotational axis, such that the optimal conditions $\sum_{j=1}^{k} R_j \cos \beta_j = 0$ and $\sum_{j=1}^{k} R_j \sin \beta_j = 0$ are satisfied. To increase the identification accuracy, four marker points are used in the calibration experiments and are denoted as $P_{01}$, $P_{02}$, $P_{03}$ and $P_{04}$, respectively. Their locations are shown in Fig. 1, where the radii $R_1 = R_3$ and $R_2 = R_4$, and the angles $\beta_3 = \pi + \beta_1$ and $\beta_4 = \pi + \beta_2$. The measurement data have been obtained using a Leica laser-tracer for the set $q_2 = \{0°, -30°, -60°, -90°, -120°, -140°\}$. The values of the identified geometrical parameters and corresponding confidence intervals are given in Table 1.

**Table 1.** Identification results for the compensator geometric parameters

|       | $L$ [mm] | $a_x$ [mm] | $a_y$ [mm] |
|-------|----------|------------|------------|
| Value | 184.72   | 685.93     | 123.30     |
| CI    | ±0.06    | ±0.70      | ±0.69      |

For the identification of manipulator elastostatic parameters, 15 measurement configurations (with 5 different values for $q_2$) were obtained based on the industry oriented performance measure (10), for which the Cartesian coordinates of the reference points ($P_1$, $P_2$ and $P_3$) were measured three times (before and after the loading). The corresponding experimental setup is illustrated in Fig. 3. The desired elastostatic parameters have been obtained using a two-step identification procedure. On the first step, the base and tool transformations have been computed. On the second step, all measurement data as well as the obtained base and tool transformations have been used for the identification of the manipulator elastostatic parameters. Corresponding numerical results are given in Table 2.

**Table 2.** Manipulator elastostatic parameters obtained using different approaches, [μrad/Nm]

|                                | $k_1$ | $k_2$ | $k_3$ | $k_4$ | $k_5$ | $k_6$ |
|--------------------------------|-------|-------|-------|-------|-------|-------|
| The results obtained in this work | 0.623 | 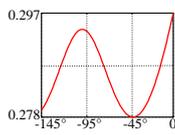 | 0.416 | 2.786 | 3.483 | 2.074 |
| (Dumas et al. 2011)            | 3.798 | 0.248 | 0.276 | 1.975 | 2.286 | 3.457 |

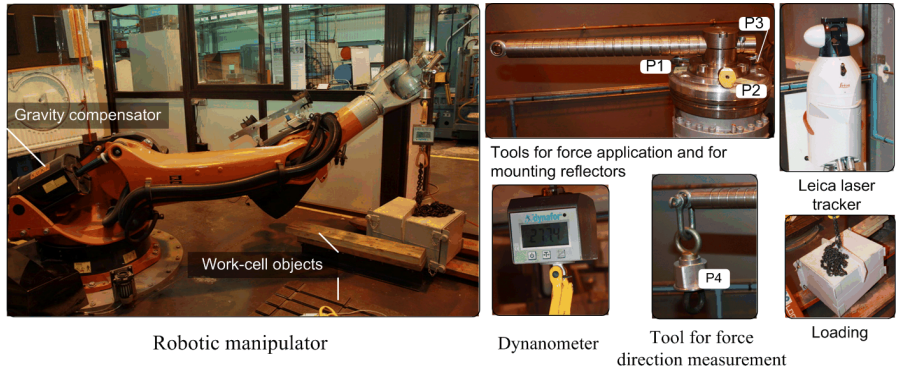

**Fig. 3** Experimental setup for manipulator elastostatic calibration

To show the advantages of the developed approach, the manipulator accuracy after calibration has been compared for two distinct plans of calibration experiments. The first one has been obtained using the industry-oriented performance measure and implements enhanced numerical routines. In this case, the manipulator was presented as a quasi-serial chain, and the calibration data were obtained using the enhanced partial pose measurements. The second plan used measurement configurations that were selected semi-intuitively, in accordance with some kinematic performance measures (Dumas et al. 2011). A relevant manipulator model corresponding to the strict serial architecture, and the calibration data were obtained using conventional full-pose measurements.

Using these two sets of calibration data, the identification yielded two slightly different sets of manipulator parameters (Table 2). Then, the obtained parameters (both sets) may be used to compute the end-effector positions for the validation configurations (that were not used in both identification routines). Comparing these results with the corresponding position measurements, it is possible to evaluate the "calibration quality" and relevant plans of the experiments.

For comparison purposes, the manipulator accuracy improvement due to elastostatic errors compensation has been studied based on the error analysis before and after compensation. Relevant results are shown in Table 3, where the maximum and RMS values of the distance-based residuals are provided. As follows from the obtained results, using the identified elastostatic parameters, it is possible to compensate 91.2% of the end-effector deflections (in average). In general, the manipulator positioning accuracy has been improved by a factor of 11.1 compare to a non-compensated robot. Compare to the previous results, the compensation efficiency has been increased by a factor of 2.4 using almost the same number of configurations, which is also referred to as the accuracy improvement factor. Hence, the above presented analysis shows the advantages of theoretical contributions presented in this work. The developed calibration technique allows us to increase essentially the manipulator positioning accuracy under external loading using a reasonable number of measurement configurations. It should be noted that the obtained elastostatic parameters can be used for elaso-dynamic analysis.

**Table 3.** The manipulator accuracy improvement after elastostatic error compensation.

| Criterion | Before compensation | After compensation | | Improvement factor | |
| --- | --- | --- | --- | --- | --- |
| | | (Dumas 2011) | [This work] | (Dumas 2011) | [This work] |
| max [mm] | 8.28 | 1.77 | 0.78 | 4.6 | 10.4 |
| RMS[mm] | 5.90 | 1.27 | 0.53 | 4.6 | 11.1 |

## 5    Conclusion

The paper deals with the accuracy improvement of a heavy industrial robot used for milling operations. It provides a sophisticated geometric/elastostatic model for quasi serial manipulators with gravity compensator and techniques for the identification of their model parameters. In order to improve the identification accuracy, design of experiments technique based on industry oriented performance measure was used. The advantages and practical significance of the proposed approach have been shown by experimental results and a comparison analysis. The improvement factor is about 2.4.

## Acknowledgements

The work presented in this paper was partially funded by the ANR (Project ANR-2010-SEGI-003-02-COROUSSO), France and FEDER ROBOTEX, France.